\documentclass[letterpaper, 10 pt, conference]{ieeeconf}  

\IEEEoverridecommandlockouts                              

\usepackage{graphics} 
\usepackage{epsfig} 
\usepackage{amsmath} 
\usepackage{amssymb}  
\usepackage[T1]{fontenc}
\usepackage{tikz}
\usepackage{pgfplots}
\pgfplotsset{compat=1.18}
\usepackage{pgf-pie}
\usepackage{caption}
\usepackage{subcaption}
\usepackage{hyperref}
\usepackage{booktabs}
\usepackage{multirow}
\usepackage[shortcuts]{extdash}
\usepackage{xcolor,colortbl}

\title{\LARGE \bf
MMDVS-LF: Multi-Modal Dynamic Vision Sensor and Eye-Tracking Dataset for Line Following 
}

\author{Felix Resch$^{*1}$, M\'{o}nika Farsang$^{*1}$, Radu Grosu$^{1}$
\thanks{F.R. and R.G. have received funding from the European Union's Horizon Europe research and innovation program with Grant Agreement No. 10039070. M.F. has received funding from the European Union’s Horizon 2020 research and innovation programme under the Marie Skłodowska-Curie grant agreement No 101034277.
}
\thanks{* denotes equal contribution}
\thanks{$^{1}$CPS, Technische Universität Wien (TU Wien), Austria}%
\thanks{E-mail of corresponding author: felix.resch@tuwien.ac.at}%
}

\newcommand{\roboracer}{\textit{roboracer} }

\begin{document}

\maketitle
\thispagestyle{empty}
\pagestyle{empty}

\begin{abstract}
Dynamic Vision Sensors (DVS) offer a unique advantage in control applications due to their high temporal resolution and asynchronous event-based data. Still, their adoption in machine learning algorithms remains limited. To address this gap and promote the development of models that leverage the specific characteristics of DVS data, we introduce the MMDVS\=/LF: Multi-Modal Dynamic Vision Sensor and Eye-Tracking Dataset for \textit{Line Following}. This comprehensive dataset is the first to integrate multiple sensor modalities, including DVS recordings and eye-tracking data from a small-scale standardized vehicle. Additionally, the dataset includes RGB video, odometry, Inertial Measurement Unit (IMU) data, and demographic data of drivers performing a \textit{Line Following}. With its diverse range of data, MMDVS\=/LF opens new opportunities for developing event-based deep learning algorithms just like the MNIST dataset did for Convolutional Neural Networks.

\end{abstract}

\section{INTRODUCTION}


An early observation during the advent of computer vision was that data-based approaches, such as Artificial Neural Networks (ANNs), need a way to obtain the data on which the approach can be developed. In the case of Convolutional Neural Networks (CNNs), these were images of digits obtained from ZIP codes of letters sent through the US postal system.
On these images, the researchers first engineered~\cite{denker1988neural} and later trained CNNs~\cite{lecun1989backpropagation} for digit recognition. These images were later published as the MNIST dataset~\cite{lecun2010mnist}, still a benchmark dataset for classification models, slowly being replaced by ImageNet~\cite{deng2009imagenet}.

This paper introduces what we hope will become a development and benchmark dataset for Dynamic Vision Sensors (DVS), an emerging sensor technology. We believe that research into new neural network models better equipped to handle the sparse, asynchronous, high-frequency nature of DVS input is a goal to work towards.

Unlike conventional camera sensors, DVSs provide an asynchronous stream of events of the form $e = (t, P, p_x, p_y)$, where $t$ denotes the timestamp, $P$ the polarity, either increasing or decreasing, and $p_x, p_y$ the coordinates of the event. These events mark changes in the per-pixel intensity and can occur at a maximum rate of a few kHz up to 1 MHz. This sparse and high-frequency scene representation is very different from the comparatively low-frequency representation even high-speed conventional frame-based cameras can provide.

Just like with ImageNet, MNIST, and its predecessors, we intend to provide a cornerstone for event-based machine learning and a benchmark dataset for developing DVS-based control models. Therefore, we introduce a multi-modal DVS dataset for a simple task in a simplified environment to encourage the development of event-based neural network theories for event-based vision.





The only existing datasets for autonomous driving with DVS sensors, DDD17~\cite{binas2017ddd17} or its successor DDD20~\cite{yuhang2020ddd20}, offer low-resolution DVS recordings and associated control inputs.
The complex scenarios recorded in those datasets make developing new ML methods challenging.
Even when only using a subset of the datasets, the environment is still very diverse and may contain observations not relevant to the task at hand.

The main challenge with these datasets is that it is difficult to determine whether a potential new ANN architecture fails to optimize due to a lack of hyperparameter tuning or a faulty novel ML theory.
We strongly believe that a reduced complexity dataset could help combat this issue. 
In our dataset we not only provide DVS data but eye-tracking data as well, which is crucial to evaluate the attention of ML models, going in the direction of explainable and trustworthy systems. 

\begin{figure}[tb]
    \centering
    \includegraphics[width=\linewidth]{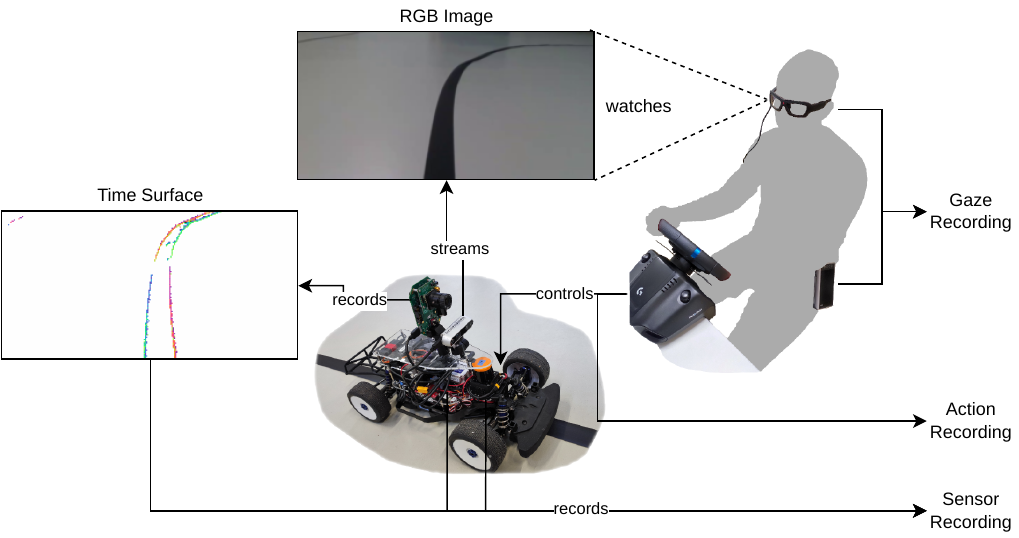}
    \caption{Recording setup for dataset recording. The human driver views the RGB stream while wearing an eye-tracking headset and controlling the vehicle remotely.}
    \label{fig:recording-setup}
    \vspace{-3ex}
\end{figure}

The key contributions of MMDVS-LF, a multi-modal DVS dataset for the \textit{Line Following} task, are recordings that contain two main modalities:

\textbf{DVS event data:} \textit{Raw events and event stream representations recorded by an event-based DVS.}

\textbf{Eye-tracking data:} \textit{The gaze of the human driver might prove crucial to evaluating the attention maps of ANNs.}

\definecolor{Gray}{gray}{0.85}
\newcolumntype{a}{>{\columncolor{Gray}}c}

\begin{table*}[tb]
    \centering
    \caption{Comparison of different DVS datasets for automotive applications. The first six datasets focus on computer-vision applications, while the others focus on control tasks. Checkmarks for the modalities indicate that data for this modality is available. Different annotation types: Manual = Manually annotated, Automatic = Algorithms were used, Implicit = Data is annotated directly from the recording.}
    \label{tab:dataset-comparison}
    \begin{tabular}{|c|l|ll|ccccc|a|r|}
        \hline
        & \textbf{Dataset} & \textbf{Task} & \textbf{Annotation} & \textbf{DVS} & \textbf{Inputs} & \textbf{IMU} & \textbf{RGB} & \textbf{Depth} & \textbf{Eye-Track.} & \textbf{Amount} \\
        \hline
        \multirow{6}{*}{{\rotatebox[origin=c]{90}{\textbf{Comp. Vision}}}} & EventVOT \cite{wang2024event} & Detection  & Manual & 1280x720 & & & \checkmark & & & 249.92GB \\
        & FELT \cite{wang2024FELTSOT} & Detection  & Manual & 346x260 & & & \checkmark & & & 664.78GB \\
        & 1 MP Automotive \cite{perot2020learning} & Detection & Automatic & 1280x720 & & & & & & 15h/3.5TB \\
        & MVSEC \cite{zhu2018multi} & Depth Est.  & Implicit & 2x346x260 & & \checkmark & 2 x Gray & \checkmark & & 186.62GB\\ 
        & DSEC \cite{gehrig2021dsec} & Depth Est.  & Implicit & 2x640x480 & & \checkmark & 2 x \checkmark & \checkmark & & 453GB\\
        & Vivid++ \cite{lee2022vivid++} (Driving) & Visual SLAM  & Implicit & 640x480 & & \checkmark & \checkmark & \checkmark & & 4:19h \\ 
        \hline
        \multirow{4}{*}{{\rotatebox[origin=c]{90}{\textbf{Control}}}} & Moeys et al. \cite{moeys2016steering} & Following  & Manual & 36x36 & & & \checkmark & & & 1:15h \\
        & DDD17 \cite{binas2017ddd17} & Driving  & Implicit & 346x260 & \checkmark & & \checkmark & & & 12:00h \\
        & DDD20 \cite{yuhang2020ddd20} & Driving  & Implicit & 346x260 & \checkmark & & \checkmark & & & 51:00h \\ \cline{2-11}
        \rule{0pt}{2ex} & \textbf{MMDVS-LF (Ours)} & Line Following  & Manual & 1280x720 & \checkmark & \checkmark & \checkmark &  & \checkmark & 1:35:52h \\ \hline
    \end{tabular}
    \vspace{-2ex}
\end{table*}

The dataset is further extended by (1)~Driving inputs, (2)~IMU measurements and (3)~RGB frames.

MMDVS-LF consists of recordings from human drivers performing the \textit{Line Following} task with \roboracer\cite{f1tenth} cars (standardized small-scale cars) in a simplified environment.
The car is equipped with an event-based visual sensor aimed at the floor as the primary sensory input.
The drivers use the RGB stream from the frame-based camera and have to input movement commands to remain on a line marked on the floor while continuously moving forward on that line.

%

\begin{figure*}
    \centering
     \begin{subfigure}[b]{0.245\linewidth}
         \centering
         \includegraphics[width=\textwidth]{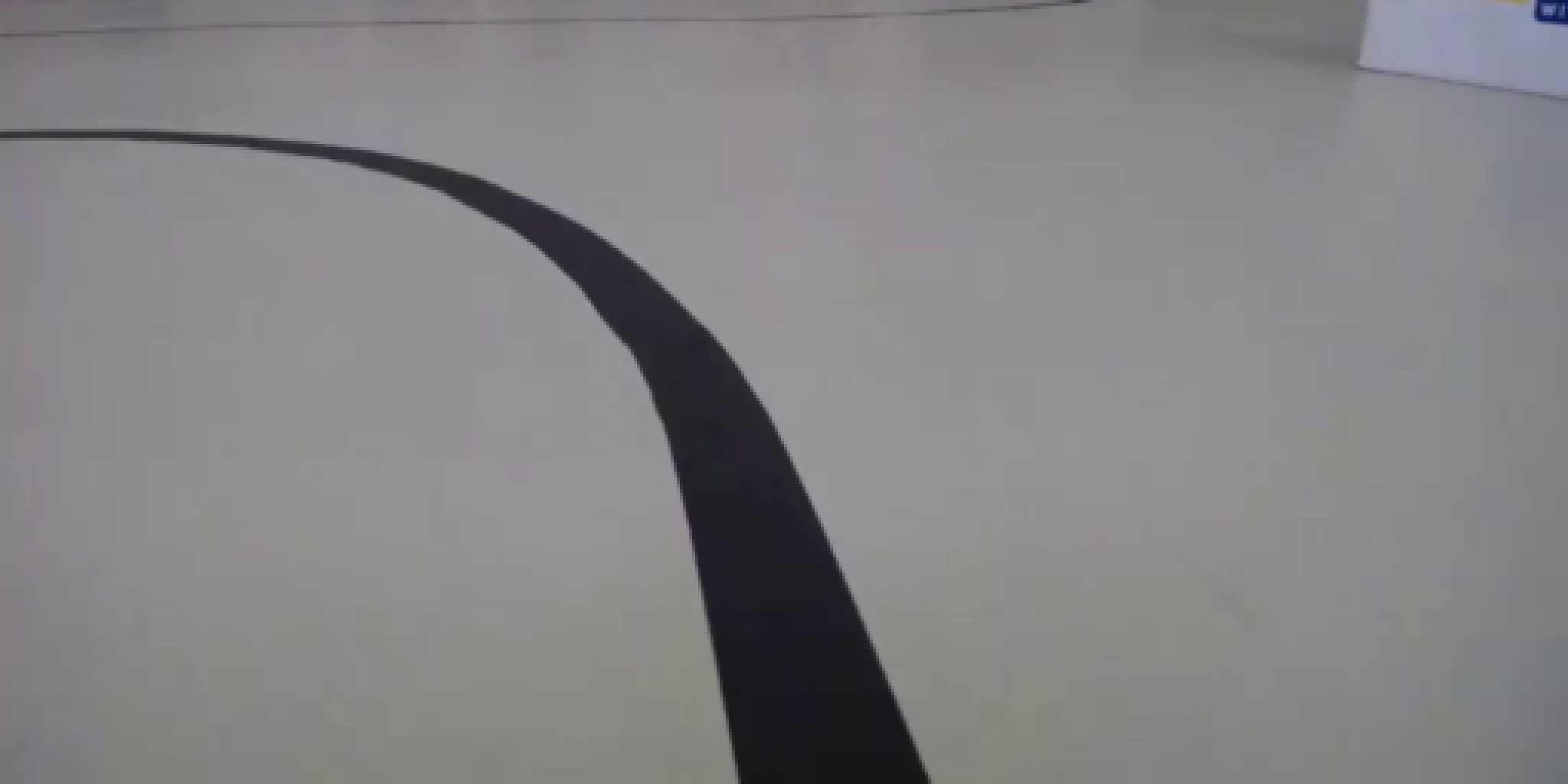}
        \caption{RGB image.}
        \label{fig:rgb}
     \end{subfigure}
     \hfill
     \begin{subfigure}[b]{0.245\linewidth}
         \centering
         {%
            \setlength{\fboxsep}{0pt}%
            \setlength{\fboxrule}{0.33pt}%
            \fbox{\includegraphics[width=\linewidth]{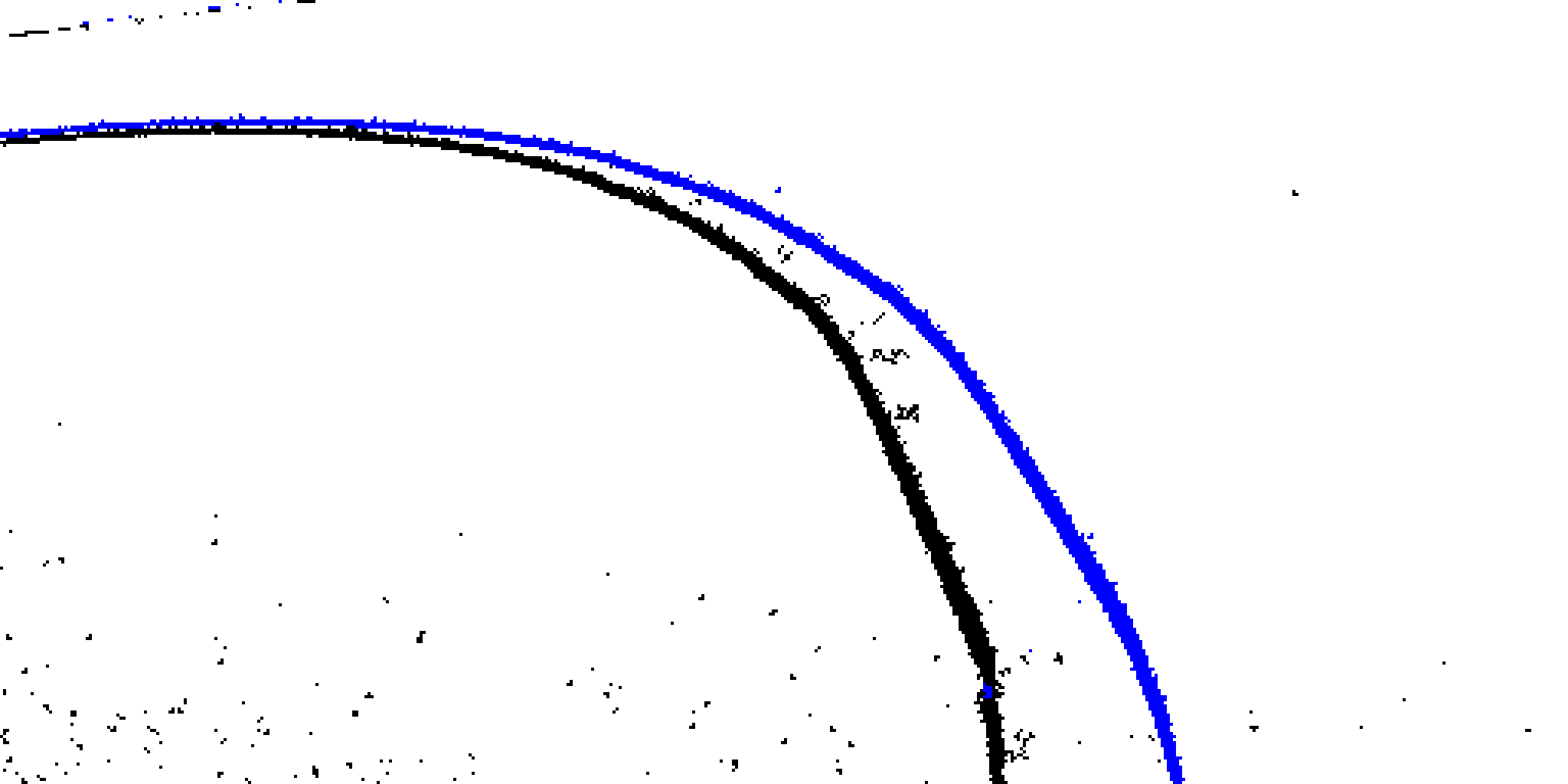}}
        }%
        \caption{Event frame.}
        \label{fig:event-frame}
     \end{subfigure}
     \hfill
     \begin{subfigure}[b]{0.245\linewidth}
         \centering
         {%
            \setlength{\fboxsep}{0pt}%
            \setlength{\fboxrule}{0.33pt}%
            \fbox{\includegraphics[width=\linewidth]{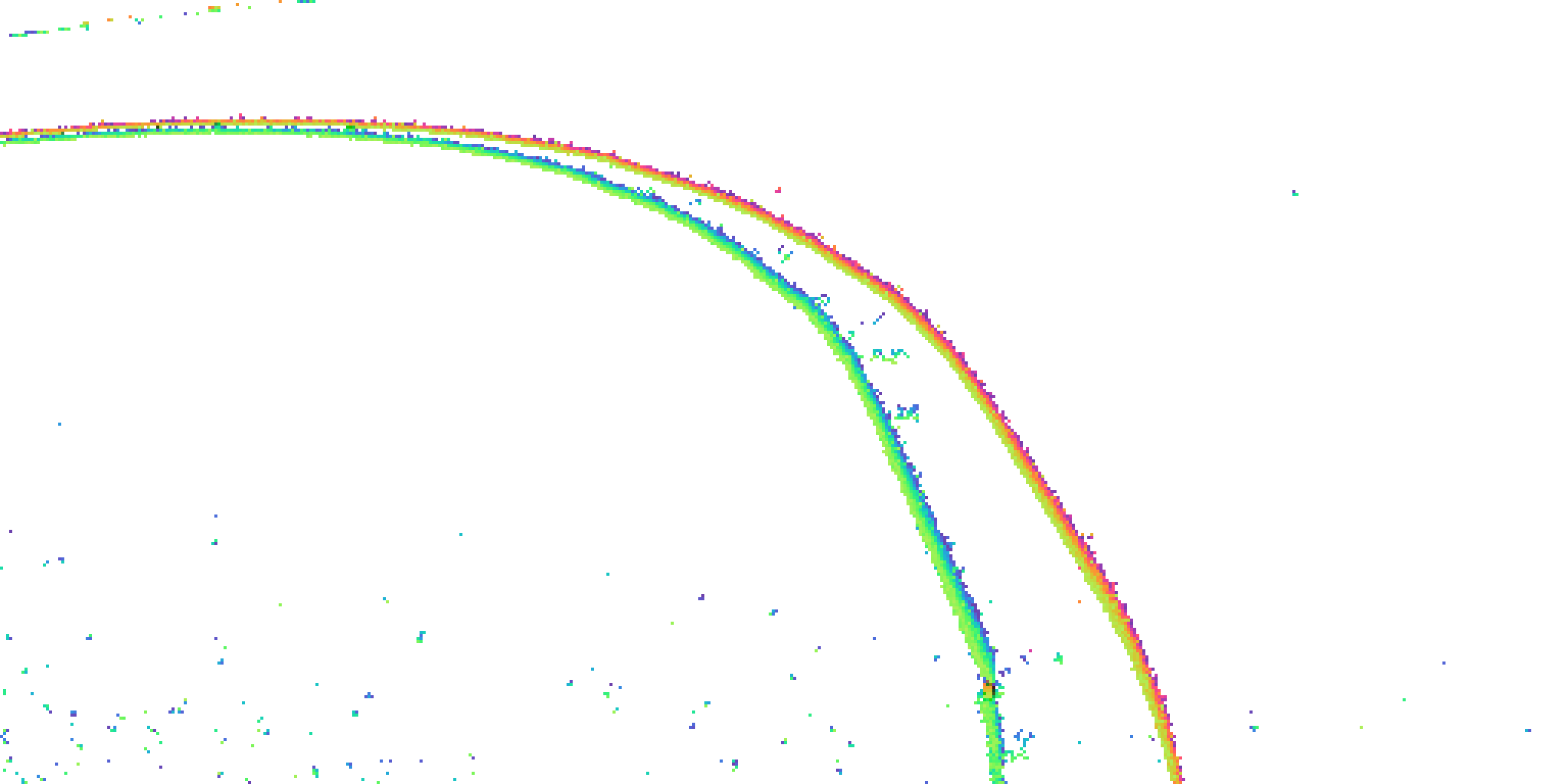}}
        }%
        \caption{Time surface.}
        \label{fig:time-surface}
     \end{subfigure}
     \hfill
     \begin{subfigure}[b]{0.245\linewidth}
         \centering
         {%
            \setlength{\fboxsep}{0pt}%
            \setlength{\fboxrule}{0.33pt}%
            \fbox{\includegraphics[width=\linewidth]{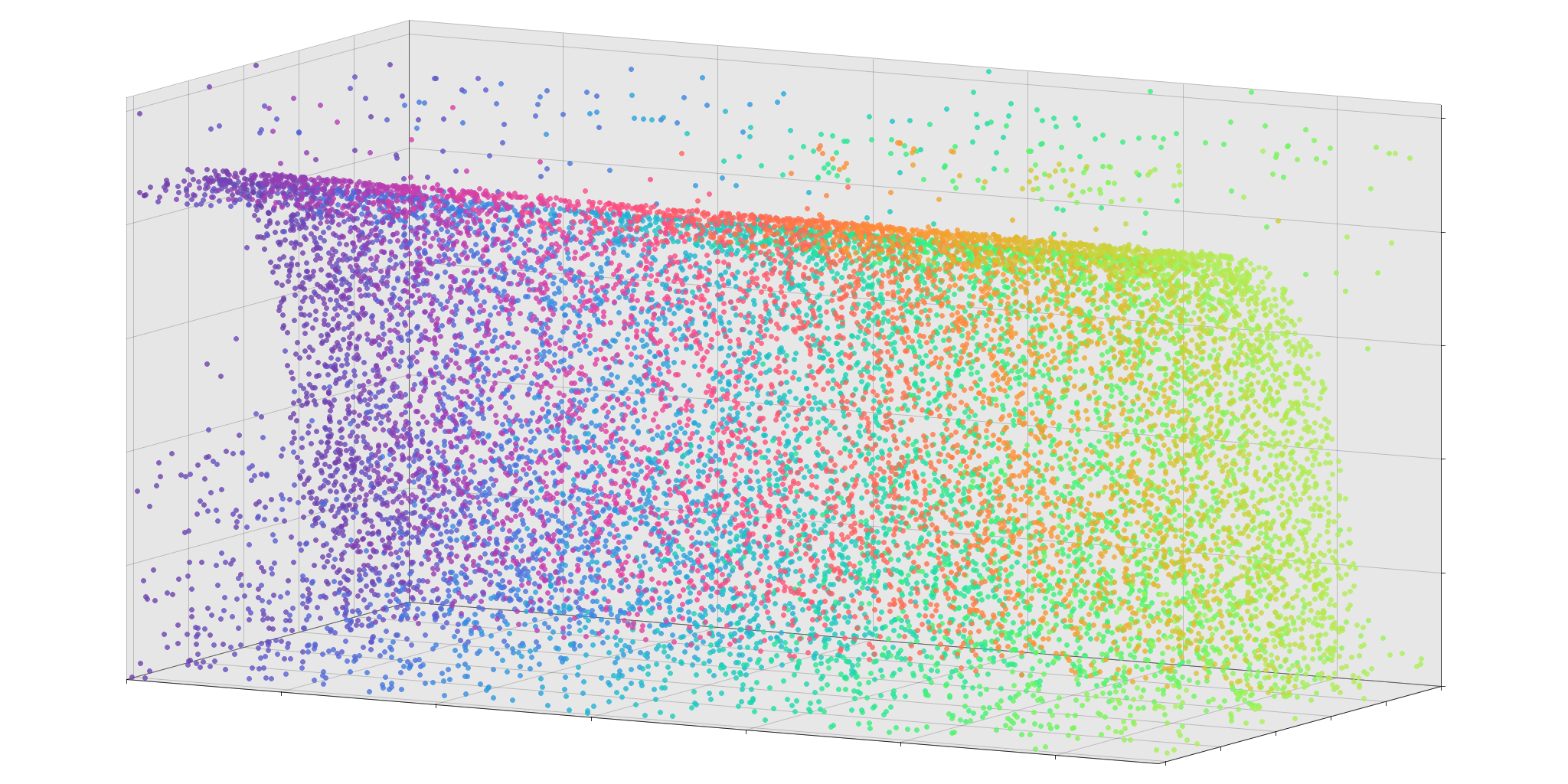}}
        }%
        \caption{Event tensor.}
        \label{fig:event-tensor}
     \end{subfigure}
    \caption{RGB frame in different corresponding DVS data representations. In the time surface and the event tensor, darker colors indicate earlier events and lighter colors later ones.}
    \label{fig:representations}
\end{figure*}

We recorded approximately 262 GB of raw data, from which we generated datasets with different resolutions and frequencies.
All generated datasets remain below 20 GB in compressed size or 50 GB if raw events are included.
Due to its compact size, MMDVS-LF is easy to use and, thanks to its simplicity, is a good choice for basic research.

This paper also demonstrates training established ANNs for a steering-prediction task based on event-based data from the dataset, for which we compare the attention maps with the eye-tracking data.

%

From the data collection and preprocessing point of view, we first give details of the recording procedure and processing pipeline for synchronizing and aligning the different modalities. Second, we describe our scaling methodology for scaling down the DVS event data.




In summary, our contributions in this paper are as follows:
\begin{itemize}
    \item MMDVS-LF, a dataset for a simple task with multiple lighting conditions, resolutions, modalities, and frequencies. To the best of our knowledge, this is the \textit{first dataset containing synchronized DVS and eye-tracking data}. 
    \item A method for collecting, synchronizing, and aligning multimodal DVS datasets.
    \item Potential use case for control application, showing how to use it with traditional CNNs in combination with Recurrent Neural Networks (RNNs) to take advantage of the temporal nature of the task. 
\end{itemize}

We provide links to the dataset files and contact information for access to the raw data at \href{https://github.com/CPS-TUWien/mmdvs}{https://github.com/CPS-TUWien/mmdvs}.

\section{RELATED WORK}

First, we review existing DVS datasets and compare them with our MMDVS-LF dataset. Then, we summarize the tasks using DVS data for deep learning control solutions in the existing literature.
Although there appears to be a larger number of computer vision datasets, the number of datasets for control tasks is limited.


\subsection{DVS Datasets for Computer Vision}

The datasets in the first section of Table \ref{tab:dataset-comparison} are designed for computer vision tasks, such as detection or visual reconstruction tasks, and do not contain driving commands. In contrast, our MMDVS-LF dataset is designed for control tasks, as it incorporates driving commands, enabling its use in tasks related to autonomous driving, such as behavioral cloning or reinforcement learning. This distinction highlights the added functionality and practical application scope provided by the MMDVS-LF dataset.

\subsection{DVS Datasets for Control Applications}
The second section of Table \ref{tab:dataset-comparison} lists datasets designed for learning control tasks.
Our MMDVS-LF dataset stands out by not only supporting control tasks but also offering synchronized data from multiple modalities, most importantly eye-tracking data.
It also includes IMU measurements and RGB frames.
These additional features provide richer context and a more comprehensive dataset, making it a valuable resource for advancing research in ML-based, trustworthy, control-oriented applications.



\subsection{Benchmark Control Tasks}\label{sec:related_benchmark}




Previous work related to control tasks using machine learning algorithms, the same way as available datasets for control, is limited. \cite{moeys2016steering} employs CNNs to predict control commands for four classes of robot movements based on DVS data. This approach restricts the robot's controllability to discrete values. A setup more similar to our work is described in \cite{maqueda2018event}, where ResNet architectures are used for event frames to predict steering angles. In contrast, we aim to explore a broader range of network architectures by employing not pure CNN-based solutions but those incorporating RNNs for sequential prediction. 


\section{SENSORS}

This section describes the novel sensor technologies we used to record the dataset.

\subsection{Dynamic Vision Sensor}

In contrast to conventional frame-based cameras, DVSs generate a stream of events in the form of $e = (t, P, p_x, p_y)$ where $t$ is the timestamp, $p_x$ and $p_y$ represent the pixel coordinates and $P$ denotes the polarity of the event.
The polarity can be either increasing or decreasing.
Once the intensity at a pixel's photosensor crosses a lower or upper per-pixel threshold, an event of the respective polarity is generated.
\textit{Events can, therefore, occur asynchronously and independently, require no periodic read-out, and achieve a high dynamic range.}\footnote{High dynamic range in frame-based photo sensors is usually achieved by taking multiple photos with different exposures. As DVSs use per-pixel thresholds, two pixels on the same chip can theoretically have an infinite dynamic range, which refers to the excitability of the individual pixel based on its luminosity.}

DVS data's asynchronous and streaming nature differs significantly from the frame-based inputs of conventional image-processing neural networks.
To address this discrepancy, typical DVS representations~\cite{gallego2022event} for ANNs try to capture the input data in a fixed-size format, such as a frame, as most architectures require fixed-sized inputs.
These representations provide formats similar to classical video frames for ANNs, allowing them to utilize established architectures by aggregating events in a specific time range.

Examples of such representations include the following.
\begin{itemize}
    \item \textit{Event Frames:} Use the polarity of the last event per pixel. (Fig.~\ref{fig:event-frame})
    \item \textit{Time Surfaces:} Use the last timestamp and polarity per pixel. (Fig.~\ref{fig:time-surface})
    \item \textit{Event Tensors:} Represent all events per pixel by including the time axis or perform fine-grained temporal aggregation. (Fig.~\ref{fig:event-tensor})
\end{itemize}


\subsection{Eye-Tracking Device}

The VPS19~\cite{vps} eye-tracking system, developed by Viewpointsystems, records a person's gaze at 60 measurements per second.
In addition to the participant's gaze, the system estimates the state of each eye, including, for example, whether a person is blinking.

A person's gaze is the position at which the foveas of both eyes are pointed.
While this area is what a person sees at high resolution, it is not necessarily the point of attention.
Visual attention in humans is generally divided into two concepts: top-down and bottom-up \cite{kastner_mechanisms_2000, johnson_attention_2004}.

Top-down attention is usually task-specific and controlled by higher cognitive functions.
Conversely, bottom-up attention is a generalized concept that reacts to motion in peripheral vision and usually leads to mental task switches if the observed visual stimulus is considered relevant enough.
After a task switch, the brain typically employs top-down attention again.
Since disturbances usually trigger a task switch and consequently focus by top-down attention, which is indicated by the gaze in visual contexts, we use the human gaze as an approximation of attention.

During the recording sessions, we asked participants to wear a VPS19 and recorded their gaze while driving. We use the gaze point projected into the RGB frame in our dataset.

\section{DATASET}

In this section, we describe the recording setup, the dataset annotation, the different formats of the MMDVS-LF dataset we provide, and statistical information.

\begin{figure*}[t]
    \centering
     \begin{subfigure}[b]{0.329\linewidth}
         \centering
         \includegraphics[width=\textwidth]{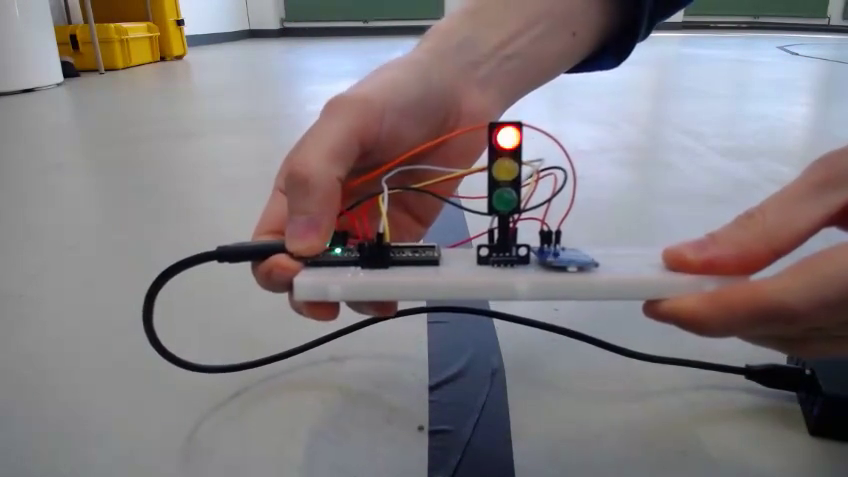}
        \caption{RGB stream.}
        \label{fig:sync-rgb}
     \end{subfigure}
     \hfill
     \begin{subfigure}[b]{0.329\linewidth}
         \centering
         \includegraphics[width=\textwidth]{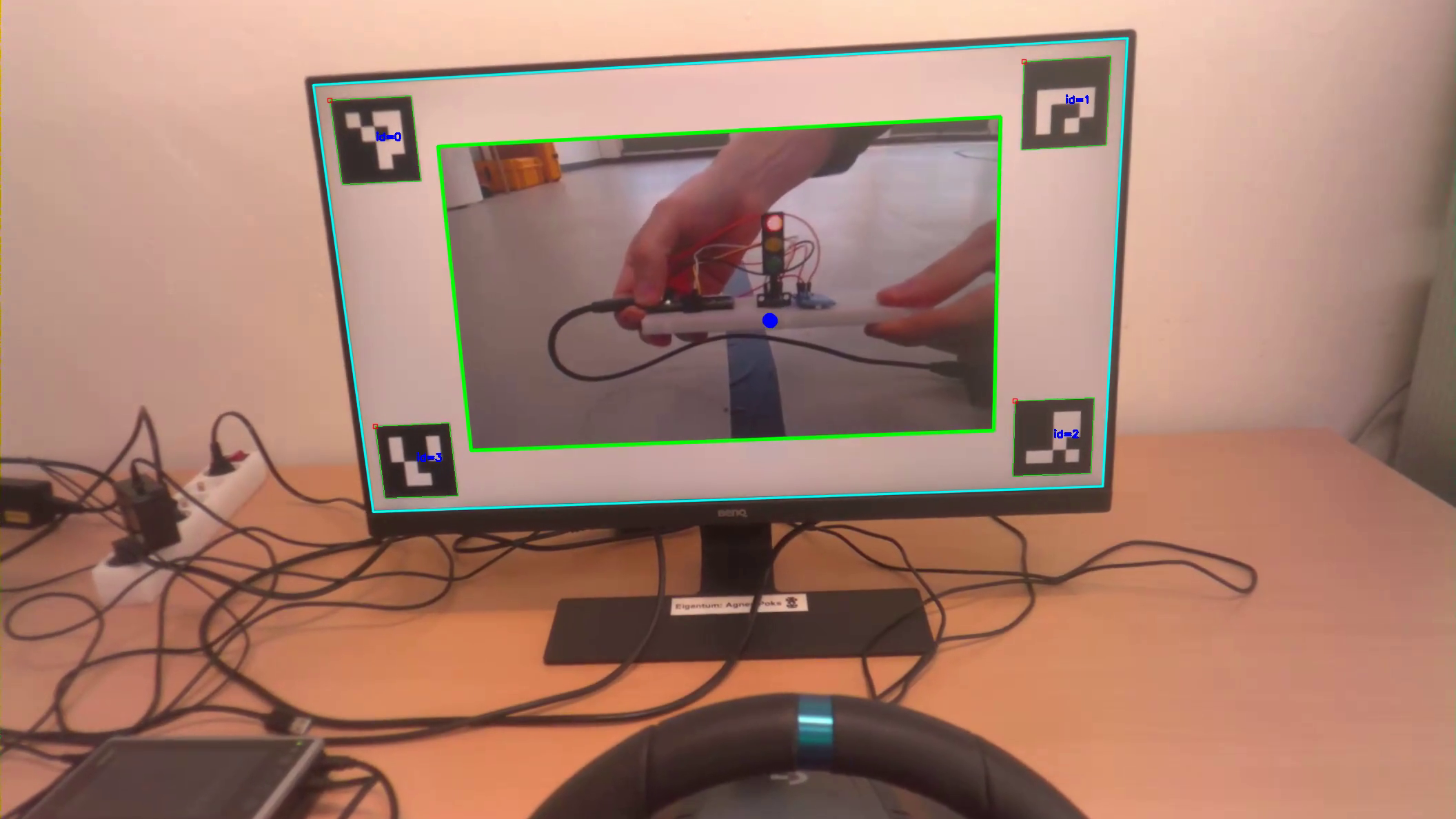}
        \caption{Annotated eye-tracking stream.}
        \label{fig:sync-vps}
     \end{subfigure}
     \hfill
     \begin{subfigure}[b]{0.329\linewidth}
         \centering
         {%
            \setlength{\fboxsep}{0pt}%
            \setlength{\fboxrule}{0.33pt}%
            \fbox{\includegraphics[width=\linewidth]{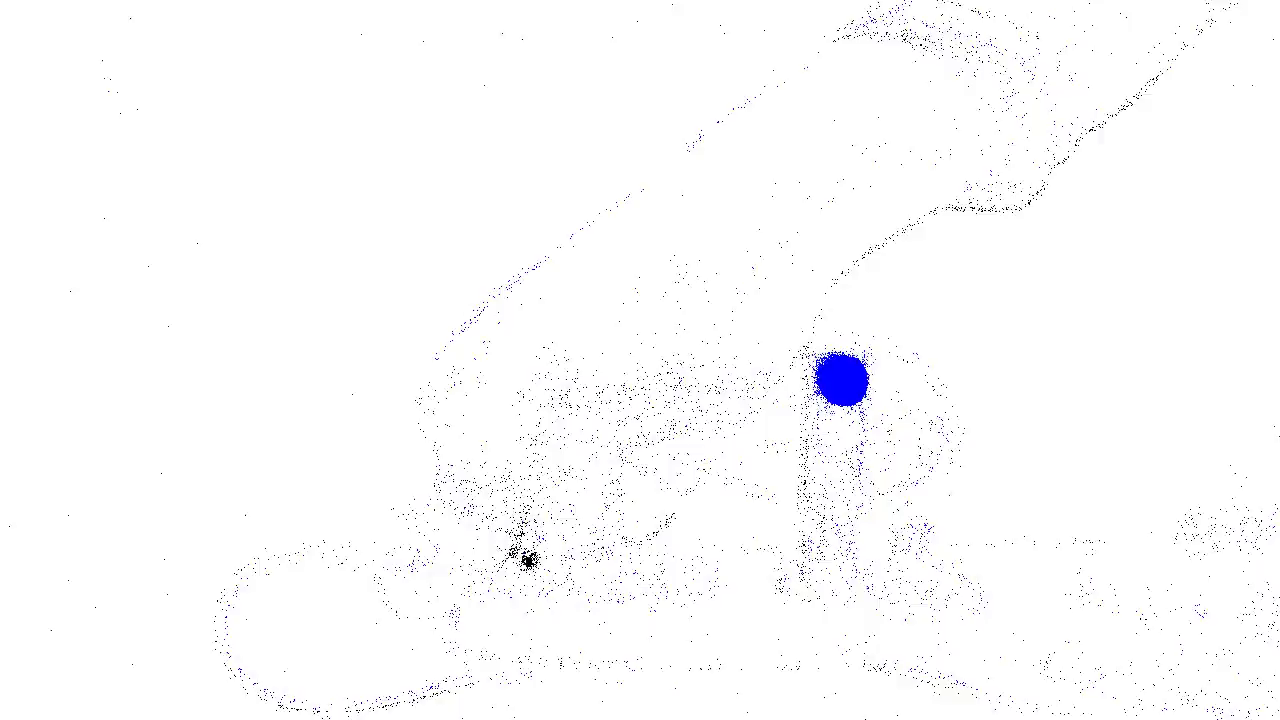}}
        }%
        \caption{DVS event frame.}
        \label{fig:sync-dvs}
     \end{subfigure}
    \caption{Temporal synchronization points between the three main temporal frames and annotated eye-tracking stream with annotated ArUco markers and RGB stream. The blue dot in the eye-tracking frame represents the gaze of the participant.}
    \label{fig:sync}
\end{figure*}

\subsection{Recording}
\label{subsec:recording}

We recorded the dataset on 1:10 scale racecars, based on the \roboracer autonomous racing cars lecture by the University of Pennsylvania~\cite{f1tenth}. 
The \roboracer cars use chassis of commercially available 1:10 model racecars and are equipped with a computing platform, motor electronics, and sensors for environment perception.
The sensors typically include a Hokuyo UST-10LX 270° 2D\=/Lidar~\cite{noauthor_ust-10lx_nodate} sensor and inertial measurement units (IMUs).
We use the Robot Operating System (ROS)~\cite{ros} to run control software for the racecar.

We mounted a Logitech C930e below a Sony Prophesee IMX636 DVS for the recording.
For mechanical reasons, the DVS was mounted slightly offset to the left, resulting in observations shifting somewhat to the right in the dataset.

The RGB video of the Logitech camera is streamed to a screen in front of a human driver, who can control the car with a steering wheel and pedals.
All other data streams, including driving commands and sensor data, are recorded on the car for later processing.
In addition to streaming the RGB video to the control station, we also recorded that stream on the car to include, for example, camera artifacts.

The remaining data is recorded with tooling from the ROS ecosystem, which includes timestamps for each recorded datum.

%

For each recording, we gave the human driver a few minutes to get comfortable with the task and the controls before recording them driving in their training direction.
After approximately five minutes, we interrupted the recording, turned the car around, and let the drivers drive in the opposite direction for another five minutes.


We also asked participants to fill out a consent form and a demographic questionnaire. This questionnaire collected their age, gender, country of origin and residence, and health details, including any chronic illnesses, visual impairments, or conditions affecting their vision. We also gathered information about their driving experience, including their length and frequency, professional or racing experience, prior experience with driving \roboracer cars, comfort level with new technology, and whether they experience motion sickness while driving.
Anonymized participants' data, including the mapping of the recordings to a driver, is available in the raw data upon request.

For the \textit{Line Following} task, we had ten participants, of whom five were born and obtained their driver's licenses in a country in Western Europe, two each in Eastern Europe and Eastern Asia, and one in Southern Europe.
We had seven male and three female participants, with 4 participants in the age bracket $25 - 29$, five in the bracket $30 - 34$, and one in the range of $35 - 39$.
One participant reported having no or less than one year of experience, one reported having 1 to 2 years, another three to five years, and the remaining seven reported having 6 to 10 years of experience.
Only one participant reported having a chronic illness, which impairs their driving skills, and $50\%$ of participants had some visual impairment.
One participant reported being a professional driver.

All the participants could perform the task without any issues, regardless of experience level.
We consciously decided to use expert and non-expert drivers for the recording, as we feared that experts might overfit on the specific track and provide more anticipatory actions rather than solely reactive ones.
For this reason, we also decided to change the driving direction after a fixed time.
Non-expert driving also leads to more upsets and subsequent recovery situations, which are more helpful for training generalizable networks.

\begin{table}[t]
    \centering
    \caption{Arrays present in a single frame file with their dimensions and a description of their contents. SIZE=\{512, 256\} refers to the resolution size of the dataset, $N$ to the number of raw events in the frame.}
    \label{tab:arrays}
    \begin{tabular}{|l|c|}
        \hline
        \textbf{Name} & \textbf{Dimension} \\ \hline
        \verb|data| & (SIZE/2, SIZE, 2) \\
        \verb|mask| & (SIZE/2, SIZE, 2) \\
        \verb|action| & (3) \\
        \verb|observation| & (20) \\
        \verb|filtered_mask| & (SIZE/2, SIZE, 2) \\
        \verb|owp_mask| & (SIZE/2, SIZE, 2) \\
        \verb|filtered_owp_mask| & (SIZE/2, SIZE, 2) \\
        \verb|raw_events| & (N, 4) \\
        \verb|human_gaze| & (2) \\
        \hline
    \end{tabular}
    \vspace{-3ex}
    
\end{table}

\subsection{Temporal and Spatial Frame Alignment}\label{subsec:alignment}

We record the data in three main temporal frames and multiple spatial frames.
To align the different data modalities, we used a modified methodology of the one we employed in previous work~\cite{resch2023attention}.

We use strobed visual impulses visible in all video streams to synchronize the three major streams, which are shown in Fig.~\ref{fig:sync}: (1) the RGB stream (Fig. \ref{fig:rgb}), (2) the eye-tracking video (Fig. \ref{fig:sync-vps}), and (3) the DVS recording (Fig. \ref{fig:sync-dvs}).
The visual impulse is timed to take at most one frame time in the two 30 FPS recording devices. 
It is also clearly visible in the DVS recording's event frame representations as a circle of increasing (blue) events.

As the recording systems might be subject to clock skew, we used six strobes at the beginning and end of each recording.
We use linear interpolation based on the synchronization points marked by the visual impulses to convert times between the different temporal frames.
All other data is in the same temporal frame as any of the frames and, at most, only offset by a constant amount.

We use a modified ArUco~\cite{garrido2016generation} placement from \cite{resch2023attention} to improve the detection of the markers.
In our modified setup, the markers have additional margins around them and also don't touch the video stream with their corners.
With this setup, we observed more reliable marker detection.
Based on the detected markers, we infer the position of the video stream in the video, determine a transformation between the eye-tracking camera frame and the displayed RGB stream, and project the gaze point from the eye-tracking to the RGB frame using OpenCV~\cite{opencv_library}.

\subsection{Annotation}\label{subsec:annotation}

We manually annotated the raw data to obtain sections of the recordings with desired behavior.
All sections where the line on the floor is visible in the bottom row of pixels in the RGB stream and where the driver manages to stay on or return to the line without losing it were considered desired behavior.
This extended acceptance leads to a broader range of recorded situations, which should also allow learning-based algorithms to learn recovering behaviors.

During some of the recordings, sunlight was visible on the floor and occasionally reached the line the participants were tasked with following.
These spots of light resulted in visual artifacts, like lens flares, in both visual sensors and caused the frame-based camera's auto-exposure to adapt to the high-intensity areas.
Although this allowed the participants to see properly while traversing a sunlit area, the RGB stream was either over- or underexposed when entering or exiting.
This led to participants driving slowly or erratically in these sections.
In the DVS recordings, the line remains visible in the sunlit areas due to the relative nature of the DVS.
As these adverse light conditions might hinder the early development of novel ANN models, we generate separate datasets without sunlit areas.

We derive the action annotations from the human drivers' driving commands and include observations from IMU and odometry.
Other sensors, such as LIDAR, were omitted from the dataset as they are irrelevant to the \textit{Line Following} task.

\subsection{Format}\label{subsec:format}

From the raw data recorded in Sec.~\ref{subsec:recording} and the annotations, we generated frame-based datasets with frequencies of 30 Hz and 100 Hz and image resolutions of 128x256 and 256x512.
The dataset with 30 Hz includes RGB images, as we use a camera with 30 FPS for recording.
We omitted the RGB images for datasets with higher frequencies to avoid using poor interpolation results.
We treat events' polarity separately for this dataset, generating two channels, one for each polarity.

To scale down the DVS data, we first crop the sensor area to a power of two and use virtual macro pixels.
Each macro pixel stores an internal state, which counts increasing and decreasing events, with events of opposing polarity canceling each other out.
Once that internal state exceeds the number of pixels in the macro pixel, the macro pixel generates an event with the respective polarity.

We generate time surfaces and event frames from the scaled-down event stream, as described in \cite{perot2020learning}.
We also provide different sets of masks, which include filters and a mode we call \verb|overwrite previous (owp)|. It removes events of opposite polarity if a more recent event occurs.
This mode performed better during initial tests with classic-control approaches, allowing algorithms to interpret only the most recent data.
We use neighborhood filtering to remove events from a frame if less than two other events occur in the adjacent pixels.

After generation, we store the dataset in compressed archives, storing each frame as \verb|.npz| file.
Storing each frame in separate files allows splitting and rearranging the datasets arbitrarily.
Table~\ref{tab:arrays} lists the arrays present in the archive and their values.
We also include index files containing continuous sections of recordings to sample continuous sections from the dataset.

All \verb|*mask| arrays represent event frames of the dataset.
The \verb|data| array might contain unfiltered arbitrary data, which must be combined with one \verb|*mask| array.
The \verb|action| consists of the steering angle and either speed or acceleration commands. The \verb|observation| array provides data from the IMU sensor, including acceleration in the (x,y,z) directions, angular velocity around these axes, and the orientation quaternion for (x,y,z,w) components. In addition to this, the \verb|observation| also includes odometry information, such as pose estimation (x,y,z), orientation quaternion (x,y,z,w), and velocity values along the (x,y,z) axes.

\subsection{Statistics}\label{subsec:statistics}

\begin{figure}[tb]
     \centering
     \begin{subfigure}[b]{0.28\linewidth}
         \centering
         \resizebox{\linewidth}{\linewidth}{%
         \begin{tikzpicture}
            \begin{axis}[
                ymin=0, ymax=145000,
                minor y tick num = 3,
                xtick distance = 0.4,
                area style, xlabel={$\alpha$ [rad]},
                label style={font=\LARGE},
                    tick label style={font=\LARGE}  
                ]
            \addplot+[ybar interval,mark=no, fill=Gray, draw=black] plot coordinates { (-0.6, 15042) (-0.5, 13916) (-0.4, 28623) (-0.3, 48687) (-0.2, 63588) (-0.1, 82483) (0, 143172) (0.1, 69035) (0.2, 57317) (0.3, 36144) (0.4, 12565) (0.5, 4640) (0.6, 0) };
            \end{axis}
        \end{tikzpicture}
        }
        \caption{Steering angle.}
        \label{fig:action-steering-angle}
     \end{subfigure}
     \hfill
     \begin{subfigure}[b]{0.28\linewidth}
         \centering
         \resizebox{\linewidth}{\linewidth}{%
         \begin{tikzpicture}
            \begin{axis}[
                ymin=0, ymax=225000,
                minor y tick num = 3,
                xtick distance = 0.4,
                area style, xlabel={$a$ [m/s\textsuperscript{2}]},
                label style={font=\LARGE},
                    tick label style={font=\LARGE}  
                ]
            \addplot+[ybar interval,mark=no, fill=Gray, draw=black] plot coordinates { (0, 8081) (0.1, 1865) (0.2, 3609) (0.3, 7871) (0.4, 25212) (0.5, 69083) (0.6, 172369) (0.7, 221238) (0.8, 65884) (0.9, 0)};
            \end{axis}
        \end{tikzpicture}
        }
        \caption{Acceleration.}
        \label{fig:action-acceleration}
     \end{subfigure}
     \hfill
     \begin{subfigure}[b]{0.28\linewidth}
         \centering
         \resizebox{\linewidth}{\linewidth}{%
         \begin{tikzpicture}
            \begin{axis}[
                ymin=0, ymax=270000,
                minor y tick num = 3,
                xtick distance = 0.4,
                area style, xlabel={$v$ [m/s]},
                label style={font=\LARGE},
                    tick label style={font=\LARGE}  
                ]
            \addplot+[ybar interval,mark=no, fill=Gray, draw=black] plot coordinates { (-0.2, 2) (0.0, 2280) (0.2, 14698) (0.4, 55961) (0.6, 108612) (0.8, 265903) (1.0, 50075) (1.2, 42133) (1.4, 24588) (1.6, 8671)  (1.8, 2019) (2.0, 270) (2.2, 0)};
            \end{axis}
        \end{tikzpicture}
        }
        \caption{Speed.}
        \label{fig:observation-speed}
     \end{subfigure}
        \caption{Distribution of driving inputs, such as steering angle and acceleration command from the human drivers and speed measured by odometry. }
        \label{fig:statistics}
        \vspace{-3ex}
\end{figure}

We generate 16 datasets with time surfaces and event frames, actions, and observations based on the different resolutions, frequencies, and the inclusion or exclusion of sections with sunlit areas, optionally including the raw events per time frame.
While the representations differ in resolution and generation frequency, the underlying data is the same, and the resulting datasets have the same action distributions.
The analysis in this section was performed on the 256x512@100Hz dataset, and sunlit sections were included.
Other datasets, especially the ones without the sunlit sections, might differ slightly. 

The generated datasets span 1:35:52 hours, including sections with sunlit areas, or 1:23:27 hours without those sections.
Depending on the frequency, this leads to datasets up to 575,213 frames for the dataset generated at 100Hz with sunlit areas. 

Figure~\ref{fig:statistics} shows the distributions of the actions taken by the human drivers during the desirable driving sections.
The steering angle's distribution is symmetric with the mean at $-0.001\ \text{rad}$, as seen in Fig.~\ref{fig:action-steering-angle}. The standard deviation is $0.224\ \text{rad}$, which is expected, as large sections of the track are straight.
As the cars were comparably heavy, no breaking was necessary, and only positive acceleration inputs (Fig.~\ref{fig:action-acceleration}) 
were recorded. The acceleration inputs
have a mean of $0.669\ \text{m/s\textsuperscript{2}}$ and a standard deviation of $0.135\ \text{m/s\textsuperscript{2}}$.
Figure~\ref{fig:observation-speed} shows that a large portion of the driving occurred with a speed in the range of $0.6-1.6\ \text{m/s}$, with a peak at $0.8-1.0\ \text{m/s}$. This peak and the fact that most other observations have a similar speed allow training neural networks to only predict for steering angle, further simplifying the network architectures.

We recorded the MMDVS-LF's data over about 10 hours, including instructions for the drivers, training, setup time, and technically required breaks, such as changing batteries.
The annotation of the dataset took approximately two weeks and generating the dataset with our tooling took approximately one week on an Intel(R) Xeon(R) Gold 6130 machine with 20 CPU cores and 64 GB of RAM.

As already pointed out, organizing and recording (in Section~\ref{subsec:recording}), aligning the frames (in Section~\ref{subsec:alignment}), and processing the dataset (in Section~\ref{subsec:annotation}~and~\ref{subsec:format}) are time-consuming and generally non-trivial tasks.
Developing the tools needed for synchronization, alignment, and processing has occurred over the last three years.
Also, organizing participants and researchers for recording sessions in a lab with a fully set up recording environment required careful planning and coordination.


\section{BENCHMARK}
DVS data offers many promising directions for deep learning research. First, we focus on the time surface representation for training machine learning models. Then, we highlight the differences in attention provided by DVS data compared to traditional RGB. Finally, we outline further potential of our MMDVS-LF dataset.




\subsection{Steering Prediction from Time Surfaces}\label{subsec:streering_from_ts}
\begin{figure}[tb]
    \centering
    \includegraphics[width=0.99\linewidth]{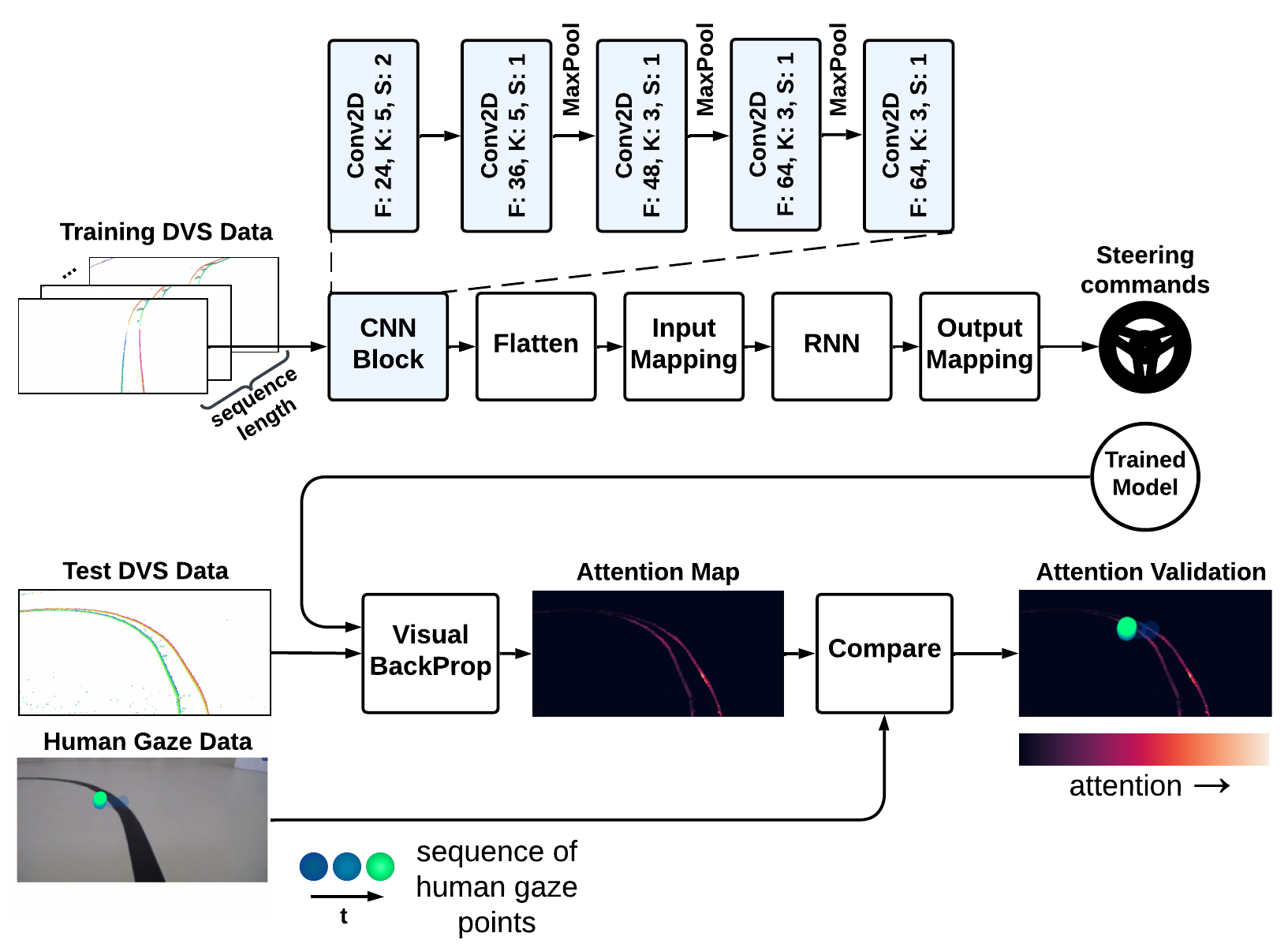}
    \caption{For benchmarking, we use the following architecture: sequences of time surface data are created and fed into our neural networks. These networks consist of a CNN Block with several convolutional and max pooling layers, followed by a flattening layer. These features are fed into a fully-connected RNN, which predicts the sequence of steering commands corresponding to the input. Before and after the RNN block, we use additional input and output mappings. For analysis, we apply the VisualBackProp method to extract the attention maps of the trained models. These are compared to the human attention from the eye-tracking data available in our dataset.}
    \label{fig:networks}
\end{figure}

Here, we present a use case for the MMDVS-LF dataset of 128x256@100Hz, where the goal is to train neural network models to predict the steering angle based on the time surface data from the DVS sensor. As pointed out in Sec.~\ref{subsec:statistics}, most of the velocity values fall into a narrow range, allowing us to simplify the task by treating the speed as constant. The pipeline is illustrated in Fig.~\ref{fig:networks}. We provide the code of a TensorFlow dataloader pipeline, and training and evaluation scripts in our \href{https://github.com/CPS-TUWien/mmdvs}{GitHub} repository. 

We trained a CNN head~\cite{Lecun1989Hand} with an RNN as a policy. As an RNN we used either a fully-connected simple RNN~\cite{chollet2015keras}, a Minimal Gated Unit (MGU)~\cite{zhou2016minimal}, a Gated Recurrent Unit (GRU)~\cite{Cho2014LearningPR}, a Long-Short Term Memory (LSTM)~\cite{Hochreiter1997Long}, a Liquid-Time Constant (LTC)~\cite{Hasani2020LiquidTN}, or a Liquid Capacitance Liquid Resistance (LRC)~\cite{farsang2024liquid} network, respectively. In these architectures, the CNN extracts visual information, while the RNN component leverages the sequential nature of the task. For configuring the CNN layers, we adapted the settings from the convolutional head used in~\cite{farsang2024learning}, which was designed to explore the task of curvature prediction based on RGB images using a combination of CNNs and bio-inspired recurrent models. This adaptation is appropriate because, at a high level, our task is similar from an ML perspective.

We compute the mean squared error (MSE) between the predicted steering angle and the ground truth values over the sequences and scale the errors by $10^{4}$ for better readability. The data is split into training, validation and test sets with a ratio of 75\%/15\%/15\%. We did hyperparameter-tuning for the learning rate in the range of $\{0.0001, 0.001, 0.01\}$. Based on their best validation loss, we train all networks using a learning rate of $0.0001$. During the training phase, we use the AdamW optimizer \cite{Loshchilov2017FixingWD} with a cosine weight decay of $10^{-6}$. We run the training for 30 epochs and save the final models with the best validation loss. They are then tested for 2,500 steps on unseen data.

\begin{figure}[tb]
    \centering
    \includegraphics[width=0.99\linewidth]{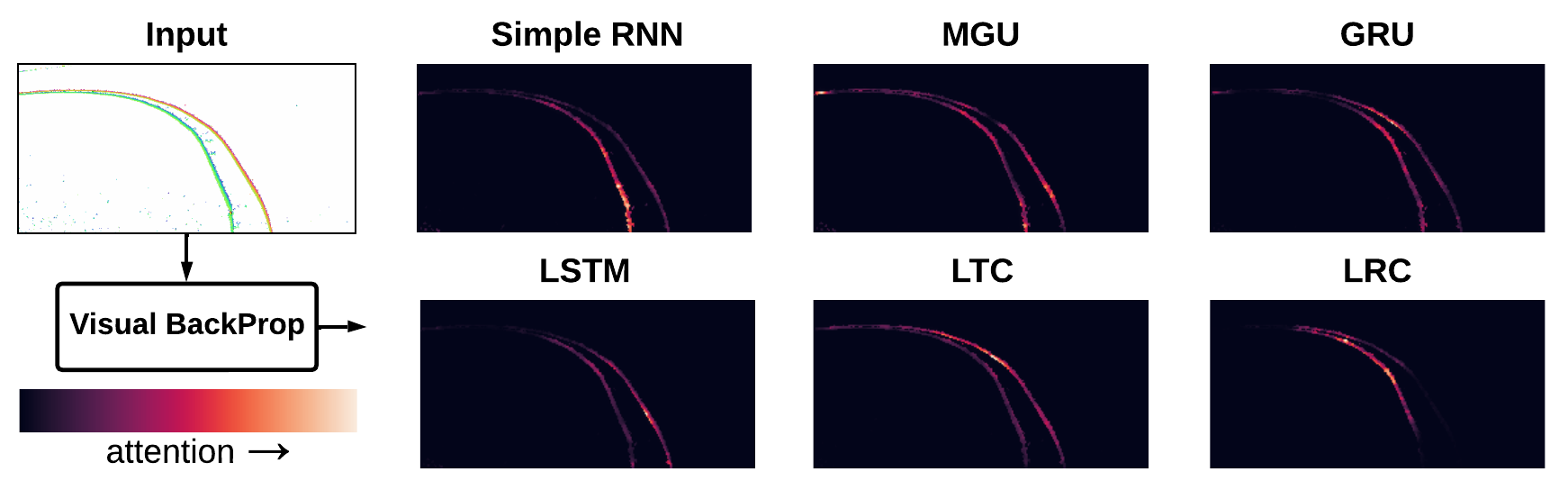}
    \caption{Attention maps of the networks, computed using the VisualBackProp algorithm, show that the models focus on different parts of the line. When compared to the human attention in Figure~\ref{fig:networks}, which displays the same frame, we found that GRU, LTC, and LRU primarily focus on the same areas as humans.}
    \label{fig:saliencies}
\end{figure}

The results of these experiments are shown in Table~\ref{tab:networks}. 
We found that all architectures were able to adapt to the task except for the CNN with Simple RNN. Our results demonstrate that more sophisticated architectures generalized better on the MMDVS-LF dataset, leading to smaller loss values. This also demonstrates that the proposed CNN head is able to extract the necessary features from the time surface, making it usable for the recurrent controllers. This is further supported by the analysis of the attention in Fig.~\ref{fig:saliencies}, which are calculated by the VisualBackProp algorithm~\cite{bojarski2016visualbackprop}, representing where the networks were focusing during decision-making. Note that all networks reduced the noise from the original input and they differ only in which part of the line ahead they prioritize.

\begin{table}[tb]
    \centering
    \caption{Training, validation, and test losses of different RNNs using the same CNN head on the MMDVS-LF  dataset. We found that CNNs in combination with MGUs, LSTMs and LRCs fit better during training and also generalize well on unseen data, which can be observed in the Test loss. Results are averaged over three seeds.}
    \begin{tabular}{|l|ccc|}
    \hline
         \textbf{Model} & \textbf{Training loss} & \textbf{Validation loss} & \textbf{Test loss}\\
         \hline
         RNN & $411.13 \pm 8.10$ & $168.13 \pm 11.98$ & $227.63 \pm 29.10$\\
         MGU & $130.48 \pm 35.58$ & $55.92 \pm 5.17$ & $\mathbf{98.74 \pm 7.51}$\\
         GRU & $146.42 \pm 41.20$ & $57.42 \pm 8.23$ & $111.24 \pm 7.57$\\
         LSTM & $143.93 \pm 33.10$& $64.56 \pm 6.45$ & $\mathbf{103.60 \pm 9.83}$\\
         LTC & $278.37 \pm 28.07$ & $100.33 \pm 14.04$& $147.69 \pm 35.77$\\
         LRC & $64.70 \pm 4.30$ & $43.48 \pm 2.78$& $\mathbf{107.29 \pm 44.46}$\\
         \hline
    \end{tabular}
    \label{tab:networks}
\end{table}

\subsection{Impact of the DVS data on Attention Maps}\label{subsec:dvs_rgb_attention}
\begin{figure}[tb]
    \centering
    \includegraphics[width=\linewidth]{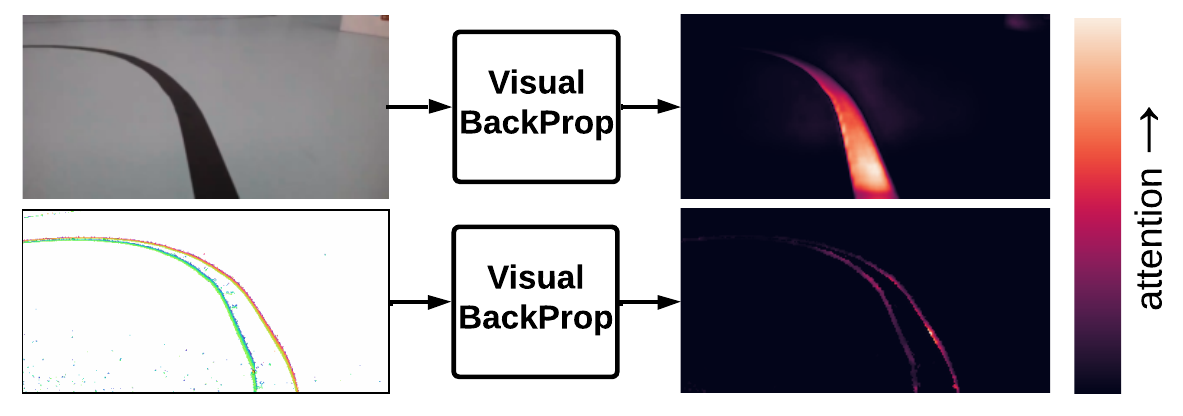}
    \caption{Demonstration of the effectiveness of DVS time surface data. The resulting attention maps show that the network focuses very precisely on the line boundaries when using DVS data, highlighting the advantage of its input representation.}
    \label{fig:rgb_dvs}
\end{figure}

Understanding how different sensor modalities influence deep learning models is crucial for improving performance in real-world applications. To address this, we investigate the impact of input representation on model attention by comparing RGB and DVS time surfaces. In this study, we trained a CNN+LSTM model separately on each data type and analyzed the resulting attention maps, which is presented in Fig.~\ref{fig:rgb_dvs}. The results show that with RGB data, the model primarily focuses on the beginning of the bold line while also slightly attending to some surrounding areas. In contrast, when using DVS data, the model eliminates the scattered area in the bottom left of the time surface representation and focuses precisely on the line boundaries, which is the most critical feature for predicting steering. This suggests that DVS input provides a highly informative minimal representation for this task, which the model can utilize effectively.

Furthermore, when comparing these findings with the different attention maps in Fig.~\ref{fig:saliencies}, we observe that the choice of the model significantly influences attention. Some models focus more effectively on the same parts of the curve as our human participants. 

To the best of our knowledge, this is the first work to analyze attention maps of DVS data, providing new insights into how event-based vision influences model interpretability and decision-making.

\subsection{Other Setups from the MMDVS Dataset}
In this section, we aim to highlight various possibilities our MMDVS-LF dataset provides, such as control tasks (regression), driver identification (classification), and other data science tasks. 
In Section~\ref{subsec:streering_from_ts}, we presented a possible setup from our MMDVS-LF dataset with a wide range of deep learning approaches. This setup can be extended by using additional available information, such as stacking RGB channels to DVS as extra input channels to the CNN part, resulting in 5 channels (3 channels from RGB, two channels from DVS) in total, and mapping other sensor information of IMU and odometry to the dense or recurrent part. One can extend the output to making sequential predictions not only on the steering angle but also on the velocity or acceleration commands. In this case, one should adapt the loss function to $L = w_{s}\text{MSE}(y_s,\hat{y}_s) + w_{v}\text{MSE}(y_v,\hat{y}_v)$ to properly scale the mean squared errors of the used commands between the ground truth labels $y_s,y_v$ and predictions $\hat{y}_s, \hat{y}_v$ by the corresponding weights~$w_s$,$w_v$, for the steering and velocity, respectively.

There are many possible training setups using the dataset, including classification tasks too, one can consider the different drivers completing the \textit{Line Following} task as class labels and use the available input data (excluding the demographic information) to make the prediction. Suppose someone aims to pursue a data science project. In that case, exploring the correlation between driving characteristics and demographic information or fault detection from the various sensor readings is possible.

By pursuing any of these directions, new ANN models could be developed to maximize the benefits of DVS data sparsity while also allowing for the integration of eye-tracking data into the training or validation pipeline.

\section{CONCLUSIONS}
We introduced MMDVS-LF, a multimodal, compact, and easy-to-use dataset primarily intended for basic research, focusing on novel deep learning solutions leveraging sparse DVS and eye-tracking data for control applications. The paper described the methods for recording experiments and constructing the dataset. We also showed several use cases of our dataset and demonstrated the power of RNNs predicting steering commands from time surface representation, and validated their attention by the eye-tracking data.

The relatively inexpensive standardized platform of \roboracer cars holds the potential to deploy end-to-end machine learning solutions on hardware, making it accessible to universities, research institutions, and the general public to test their solution developed and trained on the MMDVS-LF dataset. 







\section*{ACKNOWLEDGMENT}
We thank Mihaela-Larisa Clement, Andreas Brandstätter and Moritz Christamentl for helping with the data collection and the participants in our recordings.


\bibliographystyle{IEEEtran}
\bibliography{IEEEabrv,root}

\end{document}